\documentclass[10pt,twocolumn,letterpaper]{article}

\usepackage{cvpr}              

\usepackage{graphicx}
\usepackage{amsmath}
\usepackage{amssymb}
\usepackage{booktabs}

\usepackage[pagebackref,breaklinks,colorlinks]{hyperref}

\usepackage[capitalize]{cleveref}
\crefname{section}{Sec.}{Secs.}
\Crefname{section}{Section}{Sections}
\Crefname{table}{Table}{Tables}
\crefname{table}{Tab.}{Tabs.}

\def\cvprPaperID{3} 
\def\confName{CVPRW}
\def\confYear{2023}

\begin{document}

\title{General-Purpose Multimodal Transformer meets Remote Sensing \\ Semantic Segmentation}

\author{Nhi Kieu\\
Queensland University of Technology\\
{\tt\small v.kieu@qut.edu.au}
\and
Kien Nguyen\\
{\tt\small k.nguyentk@qut.edu.au}
\and
Sridha Sridharan\\
{\tt\small s.sridharan@qut.edu.au}
\and
Clinton Fookes\\
{\tt\small c.fookes@qut.edu.au}
}
\maketitle

\begin{abstract}
    The advent of high-resolution multispectral/hyperspectral sensors, LiDAR DSM (Digital Surface Model) information and many others has provided us with an unprecedented wealth of data for Earth Observation. Multimodal AI seeks to exploit those complementary data sources, particularly for complex tasks like semantic segmentation. While specialized architectures have been developed, they are highly complicated via significant effort in model design, and require considerable re-engineering whenever a new modality emerges. Recent trends in general-purpose multimodal networks have shown great potential to achieve state-of-the-art performance across multiple multimodal tasks with one unified architecture. In this work, we investigate the performance of PerceiverIO, one in the general-purpose multimodal family, in the remote sensing semantic segmentation domain. Our experiments reveal that this ostensibly universal network does not effectively capture the interactions between different modalities of interest in remote sensing arena. Furthermore, the network struggles with object scale variation in remote sensing images and fails to detect the presence of smaller objects such as cars from a top-down view. To address these issues, we propose a spatial and volumetric learning component, which employs 3D convolutions with an UNet configuration to encode vital local information and learn cross-modal features simultaneously, while reducing network computational burden via the cross-attention mechanism of PerceiverIO. The effectiveness of the proposed approach is validated through extensive experiments comparing it with other methods such as 2D convolution, and dual local module (\ie the combination of Conv2D $1\times1$ and Conv2D $3\times3$ inspired by UNetFormer). The proposed method significantly improves the performance of PerceiverIO, and provides competitive performance against specialized architectures like UNetFormer and SwinUNet, showing its potential to minimize network architecture engineering with a minimal compromise on the performance. Code and data will be available at https://github.com/nhikieu/SpatialVolumetricMultimodal.
\end{abstract}

\begin{figure}[t]
  \centering
  \includegraphics[width=1.0\linewidth]{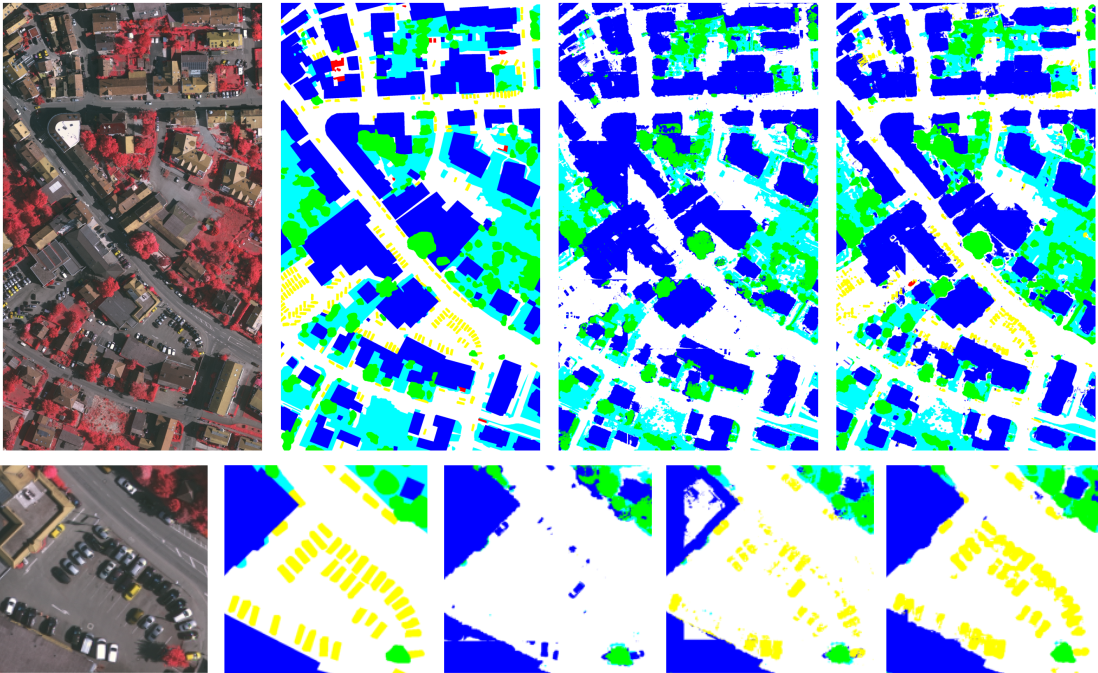}
   \caption{First row, left to right: RGIR (Red-Green-NearInfrared) image, ground truth, prediction of PerceiverIO and that of PerceiverIO with our proposed volumetric-aware module. Second row, closeup segmentation maps focusing on cars, from left to right: RGIR image, ground truth, prediction of PerceiverIO, PerceiverIO with Conv2D-based preprocessing module and that of PerceiverIO with our volumetric-aware module.}
   \label{fig:headline}
\end{figure}

\section{Introduction}
\label{sec:intro}
Semantic segmentation of remote sensing imagery refers to the task of categorizing each pixel of an image into a specific class or object to produce a dense pixel-wise segmentation map. Semantic segmentation models with good performance are crucial for the practical application of high-resolution remote-sensing images such as land cover mapping, traffic monitoring and urban management. 
However, designing remote-sensing semantic segmentation models such as UNetFormer \cite{UNetFormer} usually requires a significant amount of time, effort and domain knowledge. Moreover, adding new modalities with different structures makes the network subject to heavy re-engineering.

General-purpose transformers provide a new direction to model design by a unified architecture capable of handling different types of data in the same way. General-purpose transformers such as PerceiverIO \cite{PerceiverIO} can achieve competitive performance in multiple tasks compared to state-of-the-art domain-specific approaches.

While showing considerable promise in multimodal tasks such as learning joint representations of video, audio, and labels, the performance of these general-purpose transformers in multimodal geospatial settings has not been verified. This paper investigates the effectiveness of these techniques in multimodal settings for geospatial tasks. We apply PerceiverIO \cite{PerceiverIO} to the multimodal semantic segmentation task of very-high-resolution remote sensing and compare its performance with the state-of-the-art domain-specific approach UNetFormer \cite{UNetFormer}. Our first observation is that the PerceiverIO performs poorly on segmenting small objects such as cars. In particular, in the Vaihingen \cite{Vaihingen} and Potsdam \cite{Potsdam} datasets, PerceiverIO failed to detect cars. Our second observation is that PerceiverIO does not effectively fuse data from different modalities that are typically processed in remote sensing settings. The poor performance is firstly due to the weakness in spatial encoding, especially local information. Secondly, interactions between different modalities aren't captured to discriminating classes. We experiment multiple configurations and propose a volumetric-aware module to address these issues. \cref{fig:headline} demonstrates the effectiveness of proposed methods in detecting small objects like cars. 

\textbf{Contributions:} our main contributions in this paper are:
\begin{itemize}
    \item Contribution 1: Propose a convolution-based preprocessing component to help with small objects detection
    \item Contribution 2: Propose a volumetric-aware preprocessing component to better exploit the synergies across different modalities
\end{itemize}

The remainder of the paper is organized as follows. Section II discusses related work. Section III describes our proposed methodology. Section IV presents our datasets, experimental setup, and experimental results. The paper is concluded in Section V.

\section{RELATED WORK}
This section discusses related work in general semantic segmentation architectures, specialized semantic segmentation in remote sensing, and general-purpose multimodal architectures.

\subsection{Semantic Segmentation Architecture}
UNet \cite{UNet} is a convolutional architecture \cite{CNN_intro} that has been proven to be effective in general image semantic segmentation even though originally developed for the biomedical field. The encoder and decoder branches are independent allowing practitioners to experiment with different combinations of backbones. Hence, the idea is still widely used by the computer vision community today with more advanced backbones such as TransUNet \cite{TransUNet} and SwinUNet \cite{SwinMed}. TransUNet, for medical image segmentation, showed that Transformer can be a strong encoder while CNN remains a solid feature extractor and decoder. CNNs remain dominant in the computer vision community partially thanks to their ability of multiscale learning by progressively increasing receptive field. SwinUNet applying Swin Transformer \cite{SwinTransformer} with sliding window mechanism aims to achieve the same goal. Skip connection is an important element in UNet-like architecture, which seeks to semantically join features learnt from multiscale between encoder and decoder. However, UCTransNet \cite{UCTransNet} pointed out that there is a huge semantic gap between the encoder and decoder. Especially with a hybrid structure where the encoder and decoder are totally different in nature, the gap is even more significant. Therefore, in this work, we lean towards exploring pure transformer architecture. SegFormer \cite{SegFormer} and DC-Swin \cite{DCSwin} demonstrated that a pure attention model can extract multiscale semantic features just as well as convolutional models. In this work, we adapted SwinUNet to multimodal data to understand the performance of state-of-the-art general semantic segmentation architectures on remote sensing data.


\subsection{Specialized Architecture in Remote Sensing}
UNetFormer \cite{UNetFormer} is the current state-of-the-art architecture, specialized for remote sensing data. However, the original paper only reported results on unimodal input. Its main contribution lies in the proposal of the Feature Refinement Head and Global Local Transformer Block components in the decoder branch. In both of which, a channel path is used in conjunction with a spatial path. Even though it wasn’t explicitly explained in the paper why such design was used, we speculate that it is an attempt of capturing cross-channel features in addition to spatial features. In this work, we adapted UNetFormer to multimodal data and experimented with integrating the idea of a dual local branch into a general-purpose architecture like PerceiverIO. 

We also observe that top winners from the IEEE Data Fusion Contest 2018 (DFC2018) \cite{DFC2018_Outcome} have reported an early effort in multimodal learning. Independent branches are created for different modalities. For example, the runner-up in the contest used independent predictors for different classes. Also, heavy post-processing is required to boost performance. Since then, the potential of multimodal is gradually appreciated by the remote sensing community. Specialized architectures for tasks in geospatial settings have grown increasingly complex, pushing the boundaries of performance. The multi-stream topology is dominant within this landscape where modalities are encoded in separate branches and fused by advanced modules. While these specialized networks \cite{crossmodal_multiscale} and \cite{multimodal_graph} achieve high performance, they are largely not generalizable and will require heavy re-engineering when a new modality emerges.

\subsection{Multimodal General-Purpose Architecture}
Recently, parallel to the development of specialized multimodal architectures, more attention is given to general frameworks such as MultiMAE (Multi-modal Multi-task Masked Autoencoders) \cite{MultiMAE} and GPNA (General-Purpose Neural Architecture) with geospatial inductive bias \cite{GPNA}. These studies prove that we can use just one unified Transformer-based encoder to learn features from different modalities offering a greater degree of flexibility. PerceiverIO \cite{PerceiverIO} is an important member in this realm. It has demonstrated advantages over convolutional networks and self-attention mechanisms. Cross-attention mechanism transforms the quadratic problem into a linear problem where high resolution and high dimensional inputs can be mapped to a much smaller latent space. It also claims that the network makes little assumption about the nature of the data achieving the general-purpose goal. However, in this work, we revealed its shortcomings when it is applied to remote sensing data. Specifically, it fails to detect small objects like cars from top orthogonal inputs. In addition, it struggles to fuse information across modalities. 

\section{METHODOLOGY}
This section describes our two key contributions to address the issues when applying the general-purpose multimodal PerceiverIO to remote sensing data.

\subsection{Contribution 1}
Through our empirical experiments, it appears that the default PerceiverIO architecture with either the fixed Fourier or the learnable positional embeddings \cite{PerceiverIO,HiP} fails to perform segmentation on small objects such as cars. Even if we leverage the pretrained positional embeddings on ImageNet, the situation doesn't improve. We suspect that the model is missing crucial local information. Therefore, we first introduce an extra 2D Conv layer in the preprocessing step before feeding the inputs to the cross-attention head of the PerceiverIO. We immediately saw a huge improvement of PerceiverIO. It can detect cars, which was impossible for the default PerceiverIO. 

\begin{figure}[t]
  \centering
  \includegraphics[width=0.8\linewidth]{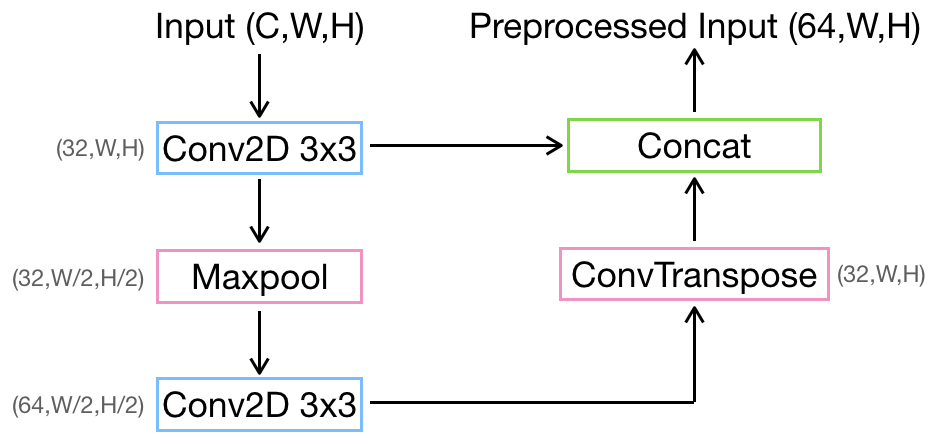}
   \caption{The Conv2D-based preprocessing module to improve local information encoding for the PerceiverIO architecture for remote sensing data.}
   \label{fig:conv2d}
\end{figure}

To put more focus on spatial information and locality, we constructed a UNet-like module (\cref{fig:conv2d}) using several 2D Conv layers to capture more local details. As expected, there is a pronounced performance boost. However, the model can only detect very bright color cars (\eg yellow, red, white) and ignore darker color cars (\eg purple, gray, black). We suspected that the prediction was greatly dependant on color channels and hasn't taken into account the complementary features from another modality which is nDSM (normalised DSM). That leads us to the second contribution.

\subsection{Contribution 2} 
\begin{figure}[t]
  \centering
  \includegraphics[width=0.8\linewidth]{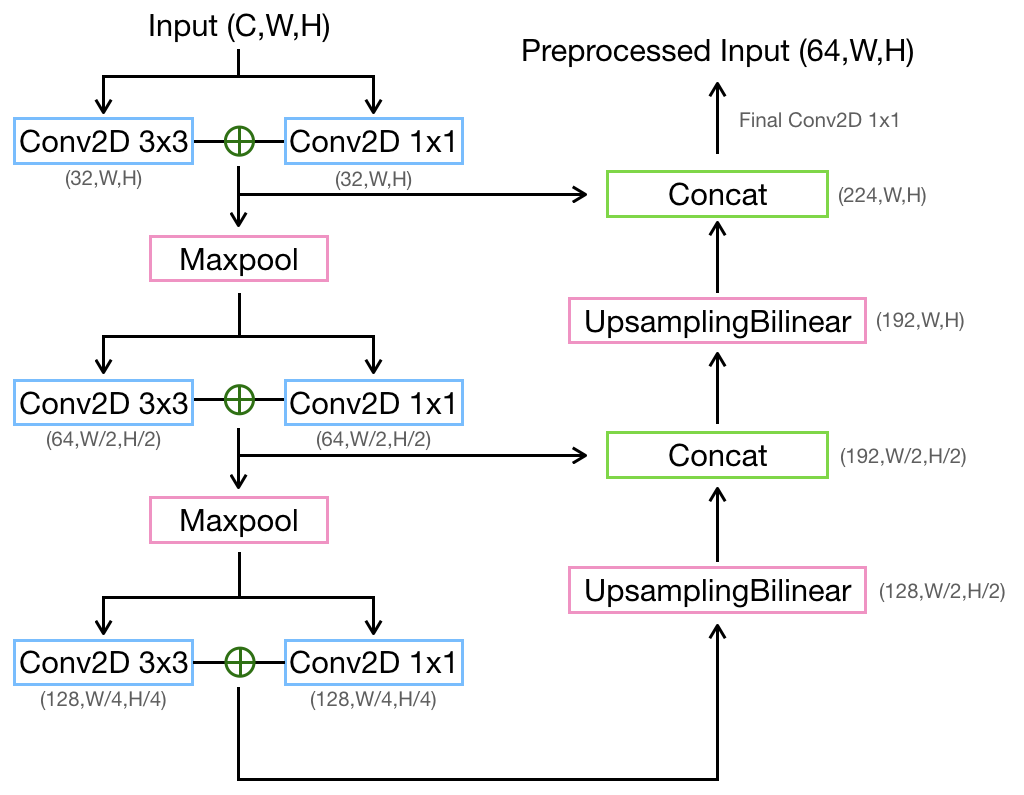}
   \caption{The Dual Local Branch preprocessing module to improve interaction among input modalities for the PerceiverIO architecture for remote sensing data.}
   \label{fig:duallocal}
\end{figure}

To improve the interaction among input modalities, \eg RGB, DSM, SAR in the remote sensing setting, we first propose a dual local branch preprocessing module. The module has two local branches: one branch uses Conv $1\times1$ and the other uses Conv $3\times3$. This is inspired by the GLTB (Global Local Transformer block) and the FRH (Feature Refinement Head) in the UNetFormer architecture \cite{UNetFormer}. In their GLTB local branch, in order to decode features, one branch uses Conv $1\times1$ and the other uses Conv $3\times3$. In their FRH, one branch is called channel path using Global Average Pool and Reduce/Expand operations, the other is named spatial path using depth-wise Conv $3\times3$. Even though these design decisions were explicitly explained by the author, it could be interpreted as an attempt to fuse spatial and channel-wise features. Inspired by this, we propose a dual local branch within our UNet-like module as shown in \cref{fig:duallocal}). 

\begin{figure}[t]
  \centering
  \includegraphics[width=1.0\linewidth]{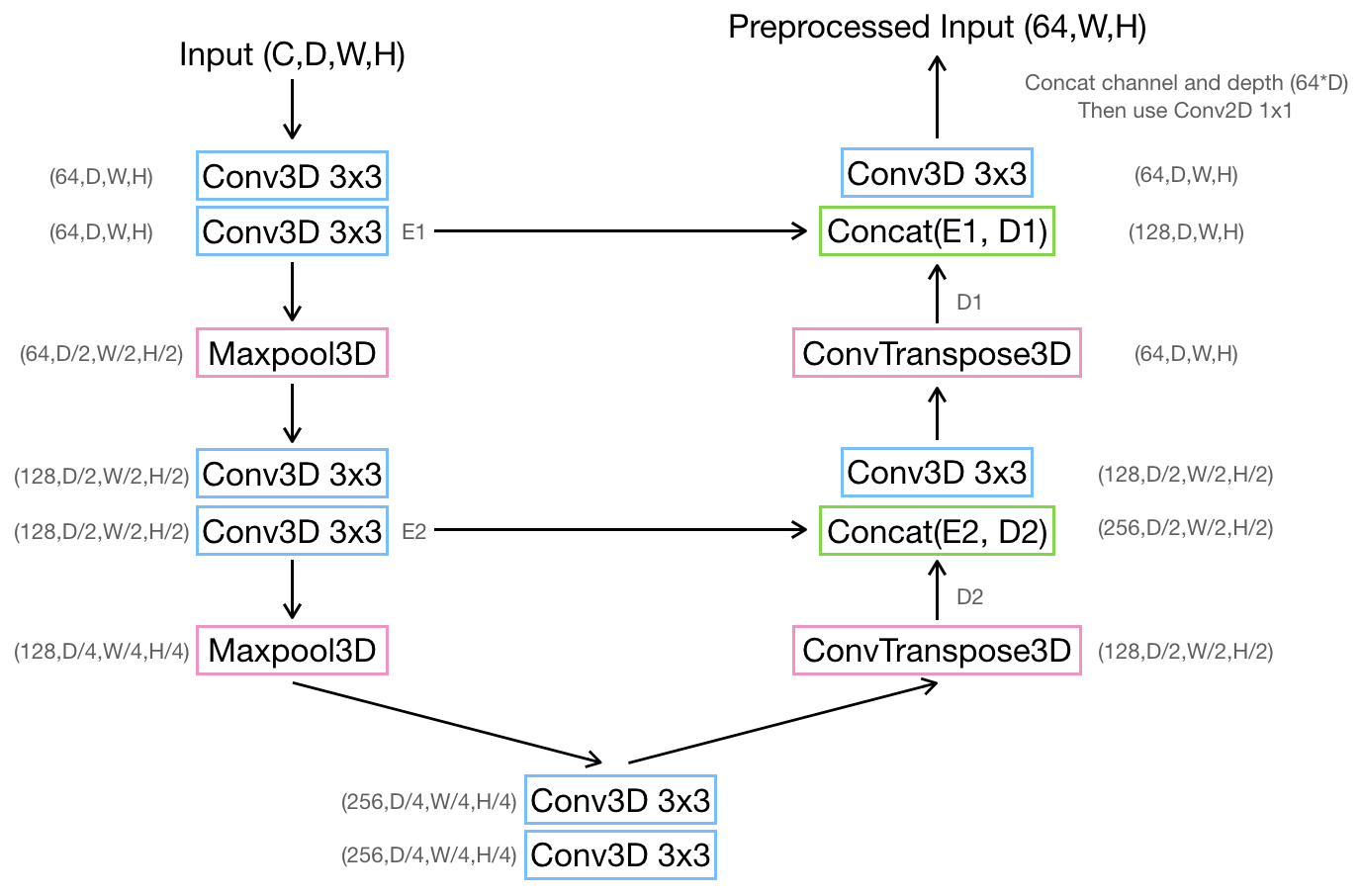}
   \caption{The Conv3D-based Volumetric-aware module to learn modality-interaction encoding for the PerceiverIO architecture for remote sensing data.}
   \label{fig:conv3D}
\end{figure}

To further improve the interaction among input modalities, we propose a Conv3D-based volumetric-aware module. The key intuition here is using 3D Convolutions will enable us to learn the interaction rather than hard coding in the network architecture. 3D Convolutional kernels can be learned to effectively fuse different input modalities for semantic segmentation. We kept the UNet-inspired design that has been work well and used 3D Conv layers to learn spatial and channel-wise features simultaneously. We observed that 3D Conv works particularly well in this situation, consolidated the volumetric nature of multimodal data even though it hasn't been widely applied. \cref{fig:conv3D} illustrates the design of our preprocessing module using 3D Conv. To ensure that the global information isn't thrown away in the preprocessing step while trying to retain local information, we use multiscale architecture in both extractor and decoder branches, which can help minimize a well-known limitation of convolution operations. In the extractor line, there are three blocks of stacked $3\times3$ 3D Conv followed by a 3D Maxpool operation (except for the final block). The number of filters increases as the component goes deeper. In the decoder line, the final representation is then upsampled twice by 3D Conv Transpose operation. After every upsampling, features from higher levels in the extractor line are concatenated and parsed through another $3\times3$ 3D Conv. Finally, channels and depth are combined and re-projected using $1\times1$ 2D Conv resulting a preprocessed input that is ready to parse through the PerceiverIO network.

\section{EXPERIMENTAL RESULTS}
This section presents our datasets, experimental setup, and experimental results.

\subsection{Datasets}
\hspace{-14px} \textbf{Vaihingen:} The Vaihingen dataset \cite{Vaihingen} from the International Society for Photogrammetry and Remote Sensing (ISPRS) contains remote sensing data of the Vaihingen region in Germany. It has two modalities: true orthophoto (TOP) and Digital Surface Model (DSM). The TOP modality has three bands RGIR: red, green, and near infrared. The DSM modality is converted from the 3D LiDAR. It contains 33 large image tiles of different sizes with a GSD of 9 cm. Dense ground truth masks are provided for training and testing. 

\hspace{-14px} \textbf{Potsdam:} The Potsdam dataset \cite{Potsdam}, also from the ISPRS, contains remote sensing data of the Potsdam region in Germany. The data set contains 38 patches of the same size, each consisting of a true orthophoto (TOP) and a DSM. The ground sampling distance of both, the TOP and the DSM, is 5 cm. Different to Vaihingen, Potsdam's TOP modality has four bands RGBIR: red, green, blue, and near infrared.

\begin{figure}[t]
  \centering
  \includegraphics[width=1.0\linewidth]{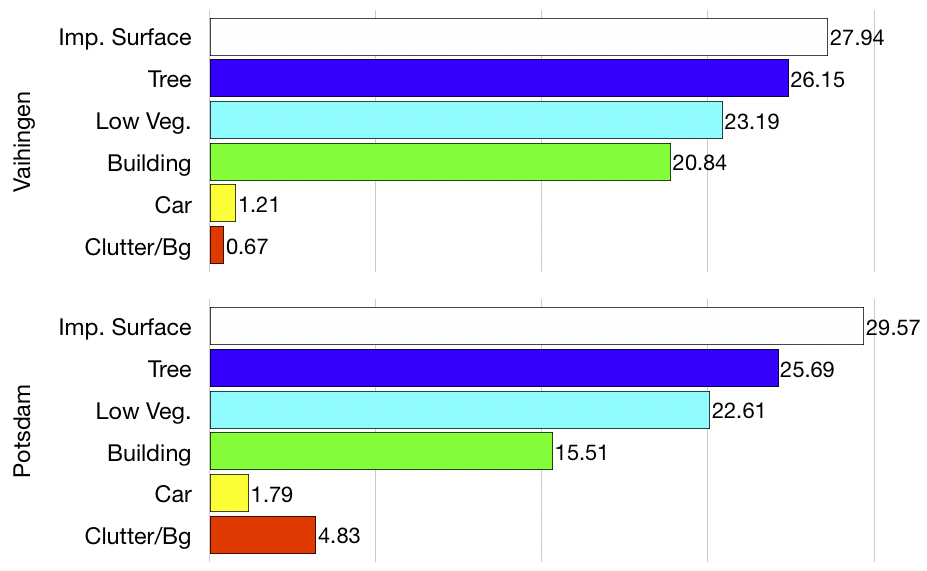}
   \caption{Class proportion of Vaihingen and Potsdam dataset. Severe class imbalance is present.}
   \label{fig:class_proportion}
\end{figure}

It's worth noting that both datasets are heavily imbalanced as shown in \cref{fig:class_proportion}, which makes it very challenging for the network to pick up already hard-to-learn small objects like cars.

\hspace{-12px} \textbf{MMFlood:} MMflood is a multimodal dataset used for flood monitoring and analysis. It includes data from Synthetic Aperture Radar (SAR - VV and VH channels), Hydrography and DEM (Digital elevation model). However, this dataset is very challenging because of two major issues: (1) More than half of the hydrography information is missing for train, (2) Severe class imbalance between flood area and background. 

\begin{figure}[t]
  \centering
  \includegraphics[width=1.0\linewidth]{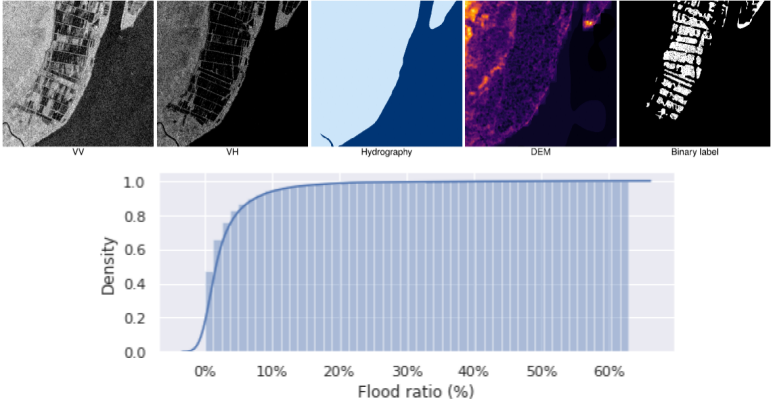}
   \caption{A sample from mmflood dataset and Cumulative distribution function of the flooded areas ratios, which demonstrate severe class imbalance issue.}
   \label{fig:mmflood_dataset}
\end{figure}

\subsection{Experimental Setup}
Selected tiles for train, validation and test are as specified on ISPRS data portal \cite{Vaihingen, Potsdam}. For training purposes, from 15 large tiles of varying dimensions provided by the Vaihingen dataset, we generated 1,620 samples of size $512\times512$. Similarly, we created 3,466 samples of size $512\times512$ for the Potsdam dataset from 22 large tiles with diverse dimensions. Specifically, tiles with the following IDs are used: (1) \textbf{Vaihingen}: \textit{Train} [1,3,5,7,11,13,15,17,21,23,26,28,32,34,37], \textit{Validate} [30], \textit{Test} [2,4,6,8,10,12,14,16,20,22,24,27,29,31,33,35,38]; (2) \textbf{Potsdam}: \textit{Train} ['2\_11', '2\_12', '3\_10', '3\_11', '3\_12', '4\_10', '4\_11', '4\_12', '5\_10', '5\_11', '5\_12', '6\_7', '6\_8', '6\_9', '6\_10', '6\_11', '6\_12', '7\_7', '7\_8', '7\_9', '7\_11', '7\_12'], \textit{Validation} ['2\_10'], \textit{Test} ['2\_13', '2\_14', '3\_13', '3\_14', '4\_13', '4\_14', '4\_15', '5\_13', '5\_14', '5\_15', '6\_13', '6\_14', '6\_15', '7\_13']

Multimodal data is introduced to selected networks by stacking modalities on top of each other. For Vaihingen dataset, the multimodal input will have a shape of (512, 512, 5), where the final dimension includes Red-Green-NearInfrared, nDSM (normalised DSM), and NDVI (Normalized difference vegetation index - derived from R-G-IR channels). Similarly, Potsdam will have the same multimodal input shape of (512, 512, 5); however, the final dimension will be the combination of R-G-B-IR and nDSM. On the other hand, to tackle class imbalance issues, we experimented with different loss functions (\cref{sec:abl}). We found a joint loss of Dice \cite{DiceLoss} and Soft Cross-entropy without class weight perform the best. This joint loss function was applied in all reported experiments. 

\begin{equation}
  L = L_{dice} + L_{ce}   
\end{equation}

\begin{equation}
L_{dice} = 1 - \frac{2}{N} \sum^N_{n=1}\sum^K_{k=1}\frac{\hat{y}^n_k y^n_k}{\hat{y}^n_k + y^n_k}
\end{equation}

\begin{equation}
L_{ce} = - \frac{1}{N} \sum^N_{n=1}\sum^K_{k=1}y^n_k log\hat{y}^n_k
\end{equation}
where $N$ is the number of samples and $K$ is the number of classes. $y^n_k$ is the one-hot encoding map of true segmentation label of sample $n$ class $k$. $\hat{y}^n_k$ is the confidence of sample $n$ belong to class $k$ (\ie corresponding softmax output from the network).

In terms of evaluation metrics, class-wise F1 score (Dice Coefficient), mIoU (mean Intersection over Union), and Average Accuracy are used. They are calculated using the following equations:
\begin{equation}
F1 = \frac{2 \times Precision \times Recall}{Precision + Recall}
\end{equation}

\begin{equation}
IoU = \frac{\text{Area of Overlap}}{\text{Area of Union}}
\end{equation}

\begin{equation}
AA = \frac{1}{C}\sum^C_{i=1}\frac{N^i_c}{N^i_a}
\end{equation}
where for each class $i$, $N^i_c$ is the number of samples classified correctly and $N^i_a$ is the total samples.

\begin{table*}[th]
\centering
\caption{Quantitative results on the Vaihingen test set}
\label{table:vaihingen}
\begin{tabular}{l l l l l l l l l}
\hline
Method & Imp.surf. & Building & Lowveg. & Tree & Car & MeanF1 & mIoU & AA\\
\hline
PerceiverIO (original) & 0.78 & 0.83 & 0.60 & 0.78 & NaN & NaN & 0.48 & 0.90 \\

PerceiverIO + 1 layer of 2DConv & 0.77 & 0.82 & 0.61 & 0.78 & 0.35 & 0.67 & 0.53 & 0.90 \\

PerceiverIO + UNet-like 2DConv & 0.80 & 0.87 & 0.68 & 0.79 & 0.42 & 0.71 & 0.58 & 0.92 \\

PerceiverIO + UNet 2DConv 3Stages & 0.79 & 0.83 & 0.61 & 0.79 & 0.36 & 0.68 & 0.54 & 0.91 \\

PerceiverIO + UNet-like 3DConv & 0.81 & \textbf{0.86} & \textbf{0.69} & \textbf{0.82} & \textbf{0.51} & \textbf{0.74} & \textbf{0.60} & 0.92 \\

PerceiverIO + Dual Local Branch & \textbf{0.82} & \textbf{0.86} & 0.65 & 0.81 & 0.45 & 0.72 & 0.58 & 0.92 \\
\hline
\end{tabular}
\end{table*}

\begin{table*}[ht!]
\centering
\caption{Quantitative results on the Vaihingen test set}
\label{table:vaihingen_1}
\renewcommand{\arraystretch}{1.5}
\begin{tabular}{l l l l l l l l l}
\hline
Method & Imp.surf. & Building & Lowveg. & Tree & Car & MeanF1 & mIoU & AA\\
\hline
PerceiverIO\_Conv3D (Ours) & 0.81 & 0.86 & 0.69 & 0.82 & 0.51 & 0.74 & 0.60 & 0.92 \\
SwinUNet & 0.84 & 0.89 & 0.74 & 0.84 & 0.60 & 0.78 & 0.65 & 0.93 \\
UNetFormer & 0.83 & 0.88 & 0.73 & 0.83 & 0.61 & 0.78 & 0.65 & 0.93 \\
\hline
\end{tabular}
\end{table*}

\begin{table*}[ht!]
\centering
\caption{Quantitative results on the Potsdam test set}
\label{table:potsdam}
\renewcommand{\arraystretch}{1.5}
\begin{tabular}{l l l l l l l l l}
\hline
Method & Imp.surf. & Building & Lowveg. & Tree & Car & MeanF1 & mIoU & AA\\
\hline
PerceiverIO\_Conv3D (Ours) & 0.87 & 0.91 & 0.73 & 0.57 & 0.73 & 0.76 & 0.63 & 0.92 \\
SwinUNet & 0.89 & 0.95 & 0.82 & 0.80 & 0.86 & 0.87 & 0.77 & 0.95 \\
UNetFormer & 0.89 & 0.95 & 0.81 & 0.78 & 0.82 & 0.85 & 0.75 & 0.95 \\
\hline
\end{tabular}
\end{table*}

\begin{table*}[ht!]
\centering
\caption{Quantitative results on the Mmflood test set}
\label{table:mmflood}
\renewcommand{\arraystretch}{1.5}
\begin{tabular}{l l l l l}
\hline
Method & Precision & Recall & IoU & F1 \\
\hline
PerceiverIO\_Conv3D (Ours) & 0.65 & 0.76 & 0.58 & 0.65 \\
SwinUNet & 0.64 & 0.78 & 0.58 & 0.65 \\
UNetFormer & 0.64 & 0.77 & 0.57 & 0.65 \\
\hline
\end{tabular}
\end{table*}

\subsection{Experimental Results}
\cref{table:vaihingen} shows that our proposed approaches result in a pronounced performance boost for PerceiverIO, especially, it resolves the problem with the car class. The results also shows that UNet is an effective architecture for feature encoding, which encodes information through multiple scales and aggregate those features. As indicated in \cref{table:vaihingen}, the last three methods, which employ a UNet-like architecture, yield superior performance. It is also worth noting here that, the UNet-like 2D convolution module can only be effective to a certain point. When we increase to 3-stage module instead of the previous two-stage module, the result is worse. 

\begin{figure}[t]
  \centering
  \includegraphics[width=1.0\linewidth]{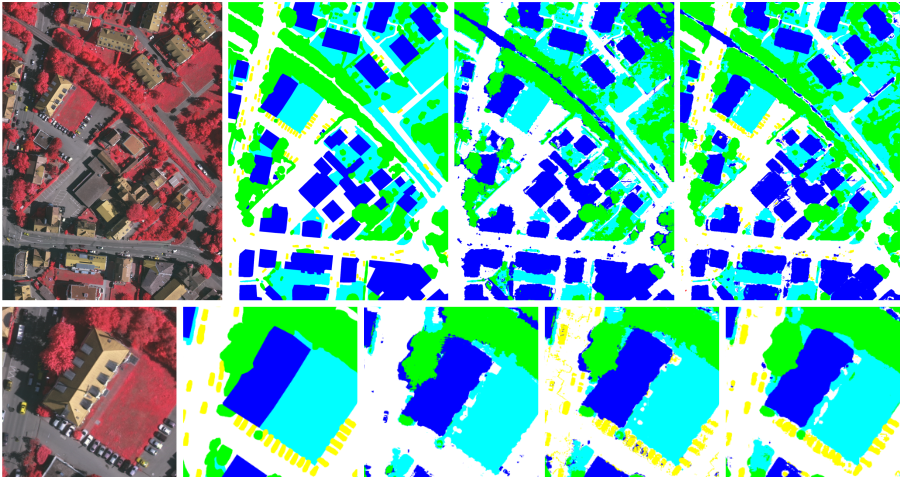}
   \caption{First row, left to right: RGIR image, ground truth, prediction of PerceiverIO and that of PerceiverIO with our proposed volumetric-aware module. Second row, closeup segmentation maps with a focus on cars, from left to right: RGIR image, ground truth, prediction of PerceiverIO, PerceiverIO with Conv2D-based preprocessing module and that of PerceiverIO with our volumetric-aware module.}
   \label{fig:vaihingen_sample}
\end{figure}

\cref{fig:vaihingen_sample} demonstrates the effectiveness of proposed methods compared to the original PerceiverIO. From the first row, it is clear that not only is the prediction of cars significantly improved, but also the overall prediction is also more realistic, devoid of obvious edge issues (\ie less misclassified pixels at the instances' boundaries). From the second row, the integration of Conv3D improves the network's ability to handle dark-colored cars and reduces prediction noise in shaded areas.

Our proposed component - local spatial and volumetric encoding - allows a multimodal, general-purpose architecture like PerceiverIO to yield highly competitive results when compared to remote sensing specialized networks like UNetFormer and segmentation specialized networks like SwinUNet on both the Potsdam and Vaihingen datasets (\cref{table:vaihingen_1} and \cref{table:potsdam}). When applied to a different dataset - MMFlood \cite{mmflood}, the three models perform very similarly; however, PerceiverIO with our proposed volumetric component and SwinUNet slightly outperform UNetFormer (\cref{table:mmflood}). MMFlood dataset is a multimodal collection of remote sensing data focused on flood monitoring and analysis. It includes data from synthetic aperture radar (SAR), and hydrography and DEM (Digital Elevation Model). However, because more than half of the hydrography modality is missing at training, it is excluded in this study. 

\begin{figure}[t]
  \centering
  \includegraphics[width=1.0\linewidth]{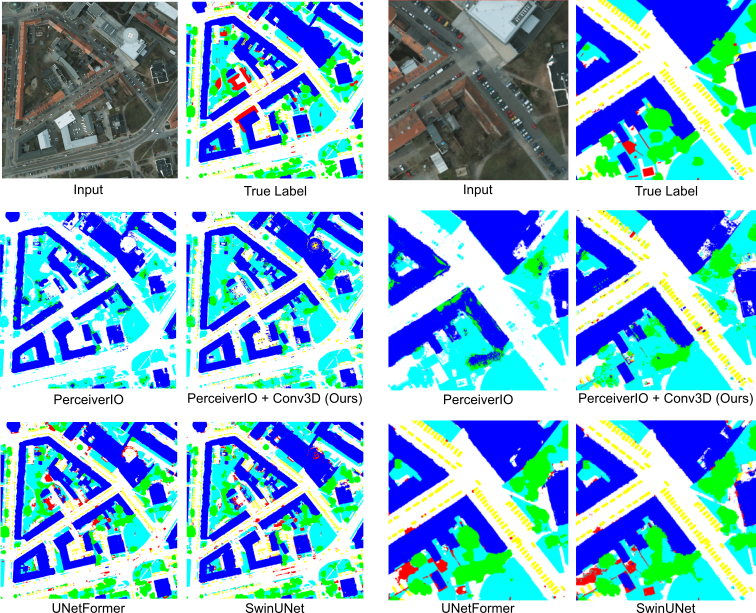}
   \caption{Qualitative demonstration on the Potsdam dataset. Our proposed method helps PerceiverIO achieve competitive results with specialized architectures like UNetFormer\cite{UNetFormer} and SwinUNet\cite{SwinMed}.}
   \label{fig:potsdam}
\end{figure}

\cref{fig:potsdam} presents several examples of semantic segmentation on the Potsdam dataset. The result is consistent with the observation in the Vaihingen dataset. The incorporation of our proposed volumetric preprocessing (UNet-inspired Conv3D) ameliorated the issue with the car class to some extent. However, we have to acknowledge that, while improved, PerceiverIO's performance is still not as precise as the specialized architectures like SwinUNet and UNetFormer, which opens opportunity for future research. Besides, a noteworthy point is SwinUNet assumes a grid-like structure for the input. Its performance hinges on a judicious choice of window size. As demonstrated in \cref{fig:potsdam}, the predictions made by SwinUNet are pixelated at the boundaries, resulting in a less smooth segmentation map compared to that generated by the PerceiverIO.

\subsection{Ablation Study}
\label{sec:abl}
To arrive at the optimal loss function, some advanced/specialized options are experimented with. They are Focal Tversky Loss \cite{FocalTversky}, Asymmetric Unified Focal Loss \cite{UnifiedFocalLoss}. However, in this case, they aren't effective because of the challenge of the object scale variation in the scene on top of the severe class imbalance issue. Assigning class weight is another option; nevertheless, it isn't easy to tune and is counter-intuitive if we want to develop a general-purpose architecture that can be applicable to different datasets. Hierarchical Perceiver (HiP) \cite{HiP} - a successor of PerceiverIO, which claims to have multiscale learning power, was explored; however, with limited data, it performed worse than PerceiverIO. We tried different positional encoding scheme suggested by HiP in an attempt to capture local features. They are Fixed Fourier 2D positional embedding, learnable positional embedding, and fine-tuned positional embeddings that were pretrained on ImageNet; however, none could resolve the issue with car detection. 

\section{Conclusion}
In this study, we proposed integrating a spatial and volumetric component into a multimodal general-purpose architecture (PerceiverIO). It helps overcome the challenge of object scale variation in severe class imbalance condition. Moreover, our experiments demonstrated the effectiveness of UNet-inspired architecture in extracting multiscale features. The baselines we used for performance comparison are specialized architectures in multimodal context (UNetFormer and SwinUNet). Our proposed method, which deploys multilayers of 3D convolutions while maintaining computing efficiency via cross-attention mechanism, provides competitive semantic segmentation results on the Vaihingen, Potsdam and MMFlood datasets. However, the development of multimodal general-purpose AI for semantic segmentation is still hindered by the expense of acquiring high-quality pixel-level annotations. In the future work, we'll introduce self-supervised and weakly-supervised learning approaches to leverage existing sparse data labels.

{\small
\bibliographystyle{ieee_fullname}
\bibliography{Paper}
}

\end{document}